\documentclass[twocolumn, switch]{article} 
\usepackage{preprint}

\usepackage{amsmath, amsthm, amssymb, amsfonts}

\usepackage[numbers,square]{natbib}
\usepackage[utf8]{inputenc}	
\usepackage[T1]{fontenc}	
\usepackage{xcolor}		
\usepackage[colorlinks = false,
            linkcolor = purple,
            urlcolor  = blue,
            citecolor = cyan,
            anchorcolor = black,
            pdfborder={0 0 0}]{hyperref}	
\usepackage{booktabs} 		
\usepackage{nicefrac}		
\usepackage{microtype}		
\usepackage{lineno}		
\usepackage{float}			
\usepackage{booktabs}       
\usepackage{caption} 
\usepackage{subcaption} 
\usepackage{multirow}

\usepackage{algorithm}
\usepackage{algpseudocode}

\usepackage{newfloat}
\DeclareFloatingEnvironment[name={Supplementary Figure}]{suppfigure}
\usepackage{sidecap}
\sidecaptionvpos{figure}{c}

\usepackage{titlesec}
\titlespacing\section{0pt}{12pt plus 3pt minus 3pt}{1pt plus 1pt minus 1pt}
\titlespacing\subsection{0pt}{10pt plus 3pt minus 3pt}{1pt plus 1pt minus 1pt}
\titlespacing\subsubsection{0pt}{8pt plus 3pt minus 3pt}{1pt plus 1pt minus 1pt}

\usepackage{tikz,xcolor,hyperref}

\definecolor{lime}{HTML}{A6CE39}
\DeclareRobustCommand{\orcidicon}{
	\begin{tikzpicture}
	\draw[lime, fill=lime] (0,0) 
	circle [radius=0.16] 
	node[white] {{\fontfamily{qag}\selectfont \tiny ID}};
	\draw[white, fill=white] (-0.0625,0.095) 
	circle [radius=0.007];
	\end{tikzpicture}
	\hspace{-2mm}
}
\foreach \x in {A, ..., Z}{\expandafter\xdef\csname orcid\x\endcsname{\noexpand\href{https://orcid.org/\csname orcidauthor\x\endcsname}
			{\noexpand\orcidicon}}
}

\title{RESTAD: REconstruction and Similarity based Transformer for time series Anomaly Detection}

\backgroundsetup{
  contents={*\textbf{Corresponding Author}: \texttt{r.ghorbani@tudelft.nl}},
  color=gray,
  scale=0.70,
  angle=0,
  position=current page.south,
  vshift=43pt  
}

\usepackage{authblk}
\usepackage{fancyhdr}

\author[1\thanks{\tt{r.ghorbani@tudelft.nl}}]{Ramin Ghorbani}
\author[1]{Marcel J.T. Reinders}
\author[1]{David M.J. Tax}

\affil[1]{Pattern Recognition Lab, Delft University of Technology, Delft, Netherlands}

\begin{document}

\twocolumn[ 
    \begin{@twocolumnfalse} 
    \maketitle

\begin{abstract}
Anomaly detection in time series data is crucial across various domains. The scarcity of labeled data for such tasks has increased the attention towards unsupervised learning methods. These approaches, often relying solely on reconstruction error, typically fail to detect subtle anomalies in complex datasets. To address this, we introduce \textit{RESTAD}, an adaptation of the Transformer model by incorporating a layer of Radial Basis Function (RBF) neurons within its architecture. This layer fits a non-parametric density in the latent representation, such that a high RBF output indicates similarity with predominantly normal training data. RESTAD integrates the RBF similarity scores with the reconstruction errors to increase sensitivity to anomalies. Our empirical evaluations demonstrate that RESTAD outperforms various established baselines across multiple benchmark datasets.

\vspace{6mm}
\end{abstract}
    \vspace{0.35cm}
    \end{@twocolumnfalse} 
] 



\section{Introduction}

Anomalies in time series data, i.e., unexpected patterns or deviations from normal behavior, can signify critical issues across various domains, from financial fraud to life-threatening health conditions. Hence, accurate anomaly detection is important. Given the rarity of anomalies and, thus, the lack of sufficient labeled data, fully supervised methods are less suited. Consequently, unsupervised learning methods have gained increasing attention~\cite{Why_Unsupervised1}. These methods do not explicitly require labeled anomaly examples, making them ideal for the detection of unknown or unexpected anomalies~\cite{Why_Unsupervised2}.

Various classic unsupervised techniques like distance-based One-Class SVM (OC-SVM)~\cite{OC-SVM} or density-based Local Outlier Factor (LOF)~\cite{LOF}, have been widely used. However, they struggle with the temporal dependencies, high dimensionality, and complex generalization demands of time series data~\cite{AD_Limit_with_TS}. Recent developments in deep learning offer promising solutions for handling these challenges~\cite{AD_TS_DeepLearning}. Architectures like Transformers and LSTMs excel at capturing temporal patterns and automatically learning hierarchical and non-linear features from time series data~\cite{Transformer_TS, AttentionAllYouNeed, LSTM_Model_Ref}.

Building on these advancements, several effective anomaly detection methods have been developed, largely focusing on the reconstruction error as a primary anomaly criterion~\cite{DCDetector, OmnyAnomaly, PatchAD}. These methods typically assess the deviations between a given input and its reconstruction to identify anomalies. The underlying assumption is that typical data will have lower reconstruction errors, whereas anomalous data will exhibit higher errors due to the unfamiliarity of the model with these patterns \cite{OmnyAnomaly, USAD, LSTM-VAE_Paper}.

\begin{figure*}[!ht]
    \centering
    \includegraphics[width=\textwidth]{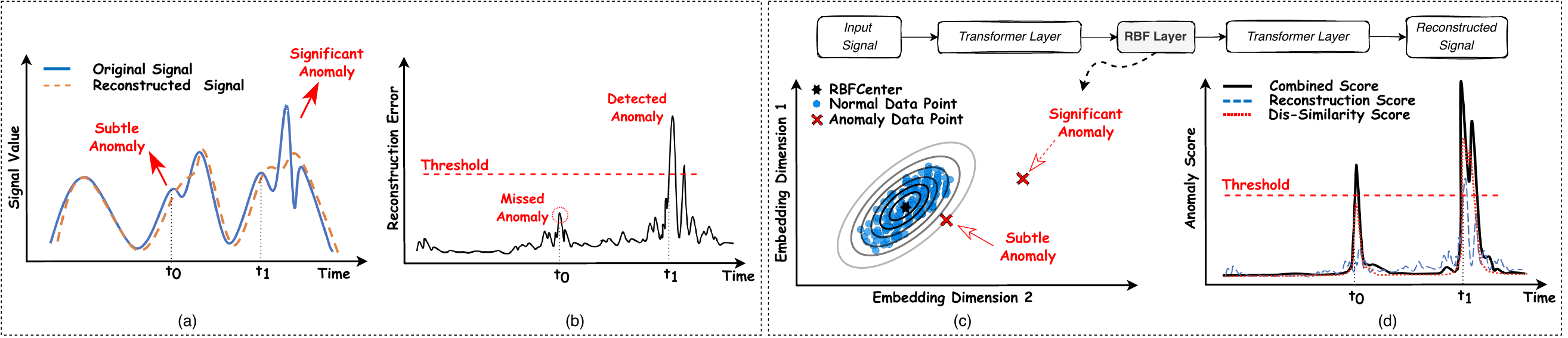}
    \vspace{-3mm}
    \caption{\emph{Comparison of traditional reconstruction and RBF-enhanced anomaly detection: a) Original signal with subtle and significant anomalies compared to its reconstruction. b) Reconstruction errors for the signals in (a), highlighting challenges in detecting subtle anomalies. c) Visualization of a model integrated with an RBF, shown via a 2D scatter plot that includes typical data, subtle and significant anomalies, and the RBF center with its influence radius, showing the RBF's ability to differentiate typical points from anomalies. d) Enhanced anomaly score using the RBF, which shows improved detection of subtle anomalies.}}
    \label{Introduction_Fig}
    \vspace{-3mm}
\end{figure*}

A major issue of using the reconstruction error for anomaly detection is over-generalization~\cite{overgeneralization_paper1}. Models fitted to capture the predominant patterns in the training data, generalize these patterns to include subtle variations as well. Therefore subtle anomalies can also be reconstructed well by these models. As a result, these anomalies are less distinguishable from typical patterns, reducing the model's detection sensitivity~\cite{overgeneralization_paper2}. This effect is depicted in Figure~\ref{Introduction_Fig}.a, where the original signal includes a subtle anomaly at time point \( {t}_0 \) and a significant anomaly at \( {t}_1 \). The reconstructed signal is a slightly smoothed version of the original signal, and by using the reconstruction error alone, the subtle anomaly is missed as the reconstruction error remains below the detection threshold, as shown in Figure~\ref{Introduction_Fig}.b.

Efforts have been made to improve unsupervised anomaly detection by adding other types of scores to the conventional reconstruction error-based anomaly scores. For instance, AnomalyTrans~\cite{AnomalyTransformer} utilizes the concept of association discrepancy, which considers the similarity of a time point with its adjacent time points. It then reweights the reconstruction error accordingly to formulate the final composite anomaly score. However, in this method a normalization is performed, which can exaggerate the discrepancy scores for normal time points when no anomalies are present, potentially leading to false positives. This can misleadingly highlight normal data points as anomalies. Although this approach is effective for identifying clear outliers, it can inadvertently misrepresent subtle normal fluctuations as anomalies.

To overcome the challenges of scoring based on reconstruction error and the limitations of the association discrepancy method, we propose combining the reconstruction error with a specialized non-linear transformation like the Radial Basis Function (RBF) kernel~\cite{RBF_Kernel}. The RBF kernel generates a similarity score that measures how close a data point is to a reference point or center, making it highly effective for anomaly detection. Anomalies, data points that deviate (are far away) from typical patterns, yield lower similarity scores with the RBF kernel, thus directly measuring how anomalous a point is. This score can effectively complement the reconstruction error and improve the sensitivity to subtle anomalies that might be overlooked by the reconstruction error. The effectiveness of combining RBF scores with reconstruction error is illustrated in Figure~\ref{Introduction_Fig}.c, where an RBF kernel is applied to typical data in the latent representation of a Transformer. By combining reconstruction error with RBF similarity scores, we create a comprehensive composite anomaly score that not only captures deviations from expected patterns but also ensures that subtle anomalies are still flagged. This composite anomaly score is shown in Figure~\ref{Introduction_Fig}.d, where the anomaly scores for both anomaly types are now above the detection threshold. 

This paper presents an adaptation of the Transformer model, chosen for its ability to capture temporal dependencies in sequential data. By integrating the RBF neurons into the Transformer architecture, we develop a model that synergistically utilizes both similarity scores and reconstruction error to compute a distinctive anomaly score. Through an extensive evaluation, we show that this new REconstruction and Similarity based Transformer for time series Anomaly Detection, RESTAD, outperforms existing baselines across a range of benchmark datasets.

\backgroundsetup{contents={}}
\vspace{-3mm}
\section{Methodology}
\label{sec:methodology}

Assume that the observed time series dataset consists of \( N \) sequences with length \( T \). Each sequence in this dataset is denoted by \( \mathcal{\boldsymbol{X}}_i = \left\{ \boldsymbol{x}_{i,t} \right\}_{t=1}^{T} \) where \( \boldsymbol{x}_{i,t} \) represents the observed time point for \( i \)-th sequence at time \( t \), having \( d \) dimensions, i.e., \( \boldsymbol{x}_{i,t} \in \mathbb{R}^{d} \). Our task is to determine if a given \( \boldsymbol{x}_{i,t} \) shows any anomalous behavior or not.

\begin{figure*}[!ht]
    \centering
    \includegraphics[width=0.78\textwidth]{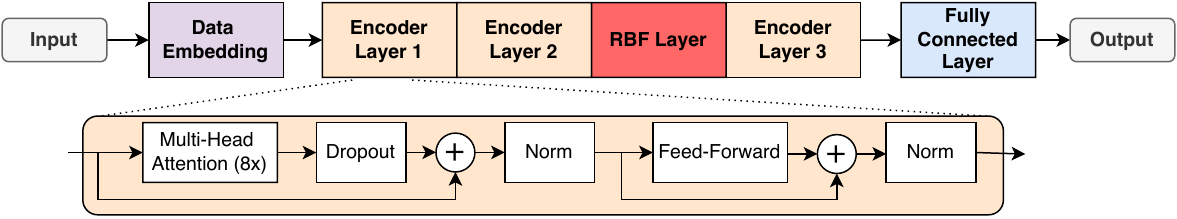}
    \caption{\emph{Overview of the proposed RESTAD model. Here, the RBF layer is added after the second encoder layer.}}
    \label{Model_Fig}
    \vspace{-1mm}
\end{figure*}
\subsection{RESTAD Framework}

In our study, we incorporate the anomaly detection mechanism into the vanilla Transformer~\cite{AttentionAllYouNeed} through a specific layer of RBF neurons, see Figure\ref{Introduction_Fig}.c. This RBF layer operates on the latent representations from the preceding layer, denoted by $\mathcal{H}_i = \left\{ \boldsymbol{h}_{i,t} \right\}_{t=1}^{T}$, where $\boldsymbol{h}_{i,t} \in \mathbb{R}^{d_h}$. This layer computes the similarity of each data point $\boldsymbol{h}_{i,t}$ to a set of learnable reference points (centers), denoted by $\mathcal{C} = \left\{ \boldsymbol{c}_{m} \right\}_{m=1}^{M}$, where $\boldsymbol{c}_{m} \in \mathbb{R}^{d_h}$. This computation results in the RBF output, $\mathcal{Z}_i = \left\{ \boldsymbol{z}_{i,t} \right\}_{t=1}^{T}$, where $\boldsymbol{z}_{i,t} \in \mathbb{R}^{M}$, which then serves as the input to subsequent layer of the model. Specifically, the RBF similarity output for each data point relative to each center is defined by:  

\begin{equation}  \label{eq:RBF}
z_{i,t}^m(\boldsymbol{h}_{i,t}, \boldsymbol{c}_m) = \exp\left(-\frac{1}{2}e^{\gamma} \|\boldsymbol{h}_{i,t} - \boldsymbol{c}_m\|^2\right)
\end{equation}

Here, the parameter \( \gamma \) controls the width of the RBF, influencing how it considers data points at varying distances from the center. This parameter is initialized and adjusted during training. Using the exponential of \( \gamma \) ensures the positivity of the scale parameter, simplifying the optimization process without enforcing a positivity constraint. 

\

\vspace{-3mm}
\noindent\textbf{Anomaly Score}: RESTAD is trained by minimizing the Mean Squared Error (MSE) to achieve accurate reconstruction. For anomaly detection, a composite anomaly score, $RESTAD_{score}$, is introduced by combining the normalized RBF similarity scores and reconstruction errors. The normalization is based on MinMax to ensure comparability. The RBF similarity score measures how closely \( \boldsymbol{x}_{i,t} \) aligns with the learned centers. A higher similarity suggests normal behavior, whereas a lower similarity (or greater distance to the RBF centers) signals anomalies. This score is derived from averaging the RBF output \(\mathbf{z}_{i,t}\) across all centers. The reconstruction error is the squared difference between the actual data \( \boldsymbol{x}_{i,t} \) and its reconstruction \(\hat{\boldsymbol{x}_{i,t}}\). The $RESTAD_{score}$ is formulated as:

\begin{equation} \label{eq:AnomalyScore}
RESTAD_{score}(\boldsymbol{x}_{i,t}) = \epsilon_r \times \epsilon_s
\end{equation}

where $ \epsilon_r = ||\boldsymbol{x}_{i,t} - \hat{\boldsymbol{x}_{i,t}}||^2$ represents the reconstruction error, and $\epsilon_s = \left(1 - \frac{1}{M} \sum_{m=1}^{M} {z}_{i,t}^m\right)$ measures dissimilarity. This combination highlights subtle anomalies characterized by both low reconstruction errors and RBF scores, as well as significant anomalies with high reconstruction errors or low RBF scores.

\
\vspace{-2mm}

\noindent\textbf{Initialization of RBF Layer Parameters}: Proper initialization of the RBF parameters, including the centers \( \boldsymbol{c} \) and scale parameter \( \gamma \), is crucial for our methodology. We explore two initialization strategies: \textit{Random} and \textit{K-means}, to assess their impact on model performance. For \textit{Random} initialization, parameters \( \boldsymbol{c} \) and \( \gamma \) are drawn from a normal distribution with zero mean and unit standard deviation. Although it is simple, it may lead to slower convergence, risk of local minima, and may not effectively represent the data distribution initially, possibly resulting in instability. In contrast, \textit{K-means} initialization uses the inherent data structure for a more representative starting point. In this approach, initially, a base model (without the integrated RBF layer) is trained to minimize the MSE of reconstruction:

\vspace{-2mm}
\begin{equation}  
\label{eq:MSE}
\mathcal{MSE} = \frac{1}{N} \sum_{i=1}^{N} \left\| \mathcal{\boldsymbol{X}}_i - \hat{\mathcal{\boldsymbol{X}}}_i \right\|^2_F
\end{equation}
\vspace{-2mm}

After achieving satisfactory reconstruction accuracy from the base model, the latent representation is extracted from the specific layer where the RBF layer is intended to subsequently be integrated. This representation is then used to initialize \( \boldsymbol{c} \) via the K-means clustering algorithm. The scale parameter \( \gamma \) is initialized using \( \tilde{\sigma}^2 \), the mean squared distance from each data point to its nearest cluster center:

\begin{equation}  
\label{eq:mean of pairwise squared distances_combined}
\tilde{\sigma}^2 = \frac{1}{N T} \sum_{i=1}^{N} \sum_{t=1}^{T} \min_m \left\| \boldsymbol{h}_{i,t} - \boldsymbol{c}_m \right\|^2, \quad \forall m \in [1, M]
\end{equation}

Here, \( \boldsymbol{h}_{i,t} \) denotes the latent representation vector of the \(i\)-th sample at the \(t\)-th time step, and \( \boldsymbol{c}_m \) is the \(m\)-th cluster center obtained from the K-means algorithm. This value, \( \tilde{\sigma}^2 \), is used to initialize \( \gamma \) as \( \gamma = \frac{1}{\tilde{\sigma}^2} \), ensuring that the RBF function has a spread informed by the average dispersion of the data points around their respective centers.

\section{Experimental Setup}
\begin{table*}[!t]
\centering
\captionof{table}{\emph{Performance metrics of baselines and RESTAD on test sets.} Initialization methods are denoted as (R) for \textit{Random} and (K) for \textit{K-means}. For all measures, a higher value indicates better anomaly detection performance.}
\label{table:Main_Results}
\fontsize{7.5pt}{7pt}\selectfont 
\resizebox{\textwidth}{!}{
\begin{tabular}{c|ccccc|ccccc|ccccc}
\toprule
Dataset & \multicolumn{5}{c}{SMD} & \multicolumn{5}{c}{MSL} & \multicolumn{5}{c}{PSM} \\
\cmidrule(lr){2-6}\cmidrule(lr){7-11}\cmidrule(lr){12-16} 
Models & \rotatebox{90}{F1-Score} & \rotatebox{90}{AUC-ROC} & \rotatebox{90}{AUC-PR} & \rotatebox{90}{VUS-ROC} & \rotatebox{90}{VUS-PR} &
\rotatebox{90}{F1-Score} & \rotatebox{90}{AUC-ROC} & \rotatebox{90}{AUC-PR} & \rotatebox{90}{VUS-ROC} & \rotatebox{90}{VUS-PR} &
\rotatebox{90}{F1-Score} & \rotatebox{90}{AUC-ROC} & \rotatebox{90}{AUC-PR} & \rotatebox{90}{VUS-ROC} & \rotatebox{90}{VUS-PR}  \\
\midrule
LSTM & $0.12$ & $0.74$ & $0.17$ & $0.79$ & $0.20$  
& $0.06$ & $0.56$ & $0.14$ & $0.63$ & $0.19$ 
& $0.11$ & $0.73$ & $0.50$ & $0.72$ & $0.51$ \\

USAD & $0.13$ & $0.63$ & $0.11$ & $0.72$ & $0.14$ 
& $0.06$ & $0.53$ & $0.14$ & $0.59$ & $0.18$ 
& $0.07$ & $0.60$ & $0.41$ & $0.61$ & $0.43$ \\

PatchAD & $0.01$ & $0.50$ & $0.04$ & $0.61$ & $0.08$ 
& $0.03$ & $0.50$ & $0.10$ & $0.57$ & $0.15$ 
& $0.02$ & $0.50$ & $0.28$ & $0.55$ & $0.33$ \\

Transformer & $0.11$ & $0.75$ & $0.19$ & $0.80$ & $0.22$ 
& $0.06$ & $0.56$ & $0.14$ & $0.63$ & $0.19$ 
& $0.13$ & $0.71$ & $0.49$ & $0.70$ & $0.50$ \\

AnomalyTrans & $0.03$ & $0.49$ & $0.04$ & $0.50$ & $0.07$ 
& $0.02$ & $0.49$ & $0.10$ & $0.52$ & $0.14$ 
& $0.02$ & $0.51$ & $0.30$ & $0.53$ & $0.34$ \\

DCDetector & $0.01$ & $0.50$ & $0.04$ & $0.51$ & $0.08$ 
& $0.02$ & $0.50$ & $0.11$ & $0.58$ & $0.15$ 
& $0.02$ & $0.50$ & $0.28$ & $0.52$ & $0.32$ \\
\midrule
\textbf{RESTAD (R)} & $\textbf{0.23}$ & $\textbf{0.78}$ & $\textbf{0.23}$ & $\textbf{0.82}$ & $\textbf{0.24}$ 
& $\textbf{0.07}$ & $\textbf{0.68}$ & $\textbf{0.18}$ & $\textbf{0.72}$ & $\textbf{0.23}$ 
& $\textbf{0.15}$ & $\textbf{0.79}$ & $\textbf{0.59}$ & $\textbf{0.76}$ & $\textbf{0.57}$ \\

\textbf{RESTAD (K)} & $\textbf{0.20}$ & $\textbf{0.79}$ & $\textbf{0.24}$ & $\textbf{0.83}$ & $\textbf{0.25}$ 
& $\textbf{0.07}$ & $\textbf{0.66}$ & $\textbf{0.18}$ & $\textbf{0.71}$ & $\textbf{0.23}$ 
& $\textbf{0.14}$ & $\textbf{0.79}$ & $\textbf{0.57}$ & $\textbf{0.76}$ & $\textbf{0.56}$ \\

\bottomrule
\end{tabular}
}
\vspace{-3mm}
\end{table*}

\subsection{Datasets and Preprocessing}
\setcounter{footnote}{0}
We use three public widely used benchmark datasets for our experiments: 1) Server Machine Dataset (SMD)~\cite{OmnyAnomaly}, 2) Mars Science Laboratory (MSL) Rover~\cite{LSTM_Model_Ref}, and 3) Pooled Server Metrics (PSM)~\cite{PSM_Dataset}. Further information on each dataset is available in our code repository\footnote{\url{https://github.com/Raminghorbanii/RESTAD}}.

Data preprocessing involves normalizing each feature to zero mean and unit variance across the time dimension. Subsequently, the normalized signal is segmented into non-overlapped sliding windows~\cite{Sliding_Window_Method} with a fixed length of 100 data points, a common setting based on previous related works~\cite{AnomalyTransformer, DCDetector}.

\subsection{Implementation}
\textbf{RESTAD Model}: The RESTAD model is an adaptation of a vanilla Transformer, incorporating an RBF kernel layer as detailed in Figure~\ref{Model_Fig}. It includes a DataEmbedding module that combines both token and positional embeddings, followed by an encoder with three layers. Each layer includes a multi-head self-attention mechanism and feed-forward networks. The model has a latent dimension of 32, an intermediate feed-forward network layer with a dimension of 128, and 8 attention heads. The RBF layer is placed after the second encoder layer (other placements are also possible, see section~\ref{se:ablation}). Optimization is performed using the ADAM optimizer, and hyperparameters are determined through systematic search to optimize reconstruction task performance. Additional hyperparameter details are available in our code repository\footnotemark[1].

\

\noindent\textbf{Evaluation}: Anomaly scores (Eq. \ref{eq:AnomalyScore}) exceeding a threshold $\delta$ are identified as anomalies. Performance is evaluated using the F1-score for threshold-dependent evaluation. Here, we follow \cite{AnomalyTransformer} by setting $\delta$ to label a predefined proportion of data points as anomalies (0.5\% for SMD, 1\% for others).  For threshold-independent analysis, we use AUC-ROC, AUC-PR, VUS-ROC, and VUS-PR metrics~\cite{VUS_Evaluation}. We exclude the point-adjustment method~\cite{PA_1} due to its overestimation~\cite{PA_K}. Our model is compared against baselines and state-of-the-arts models: LSTM~\cite{LSTM_Model_Ref}, vanilla Transformer~\cite{AnomalyTransformer}, USAD~\cite{USAD}, PatchAD~\cite{PatchAD}, AnomalyTrans~\cite{AnomalyTransformer}, and DCdetector~\cite{DCDetector}.

\section{Results}
\label{main_results}
Our empirical results, as detailed in Table~\ref{table:Main_Results}, highlight the effectiveness of the RESTAD for anomaly detection. RESTAD outperforms all baseline models across the benchmark datasets and evaluation metrics, regardless of the RBF initialization strategy. While there are slight performance differences between initialization methods, these variations are not significant enough to establish the superiority of one method over another.

\begin{figure}[!h]
\centering

\includegraphics[width=0.99\linewidth]{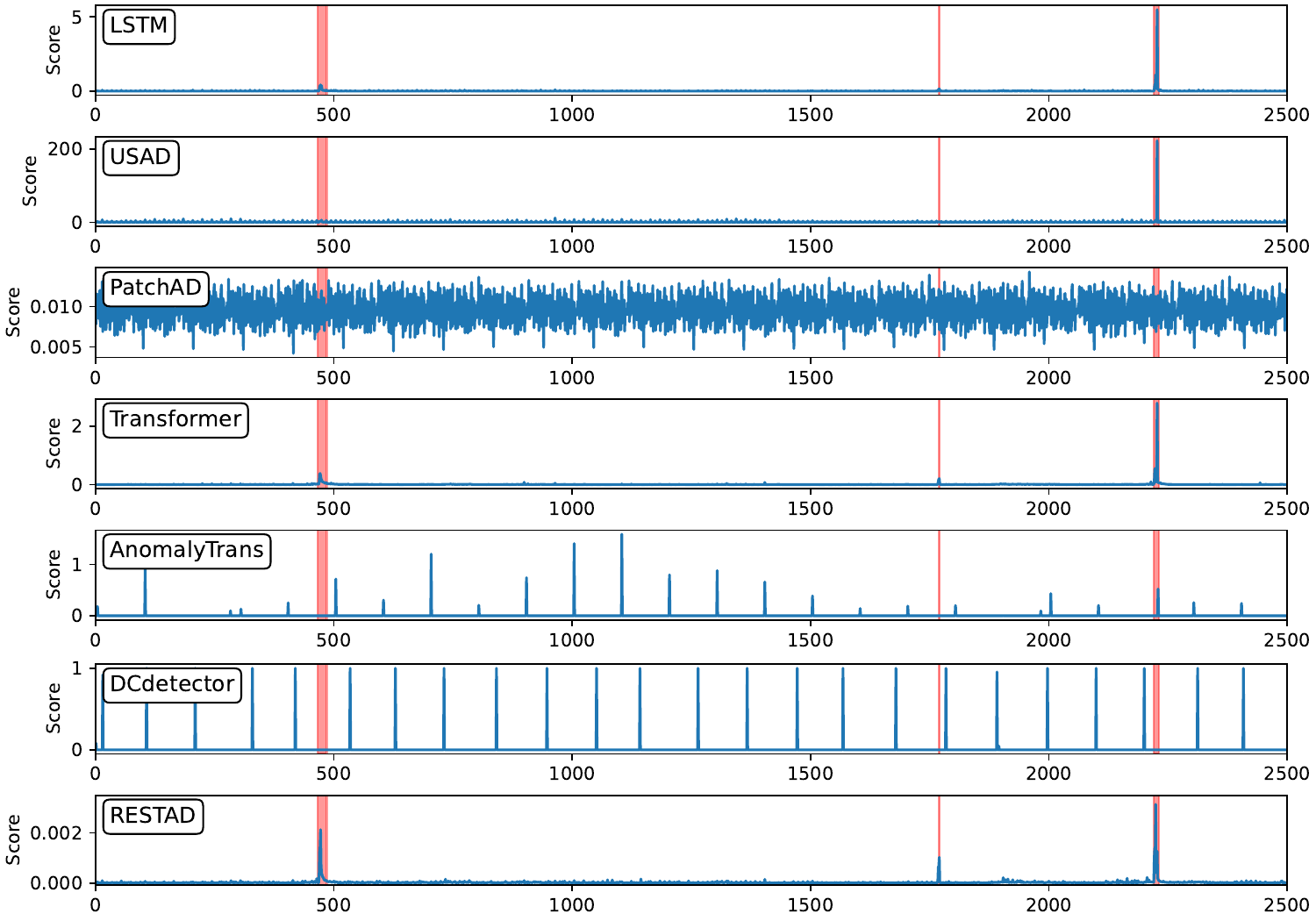}
\vspace{-1mm}
\captionof{figure}{\textit{Anomaly scores of different models for a segment of SMD dataset. The highlighted regions in red indicate the true anomaly periods (labeled by an expert).}}
\label{fig:ReSit_AScores}
\end{figure}

To visually show detection differences, Figure \ref{fig:ReSit_AScores} displays anomaly scores for a short segment of the SMD dataset. PatchAD, DCdetector, and AnomalyTrans models reveal many false detections, with DCdetector showing a pattern of repeated false positives and PatchAD resembling random scoring. LSTM, USAD, and Transformer models miss some anomalies or detect them weakly; for example, the first anomaly area is undetected by USAD, and only weakly detected by LSTM and Transformer. In contrast, the RESTAD model demonstrates robust detection, effectively identifying all anomaly sections.

\subsection{Ablation Analysis}
\label{se:ablation}
The ablation experiments are based on the RBF layer with random initialization. This decision is based on our findings that random initialization is as effective as the K-means strategy (see Table~\ref{table:Main_Results}), while offering greater simplicity and computational efficiency.

\

\noindent\textbf{Anomaly Score Criterion}: Table \ref{tab:Scores_Comparison} highlights the impact of integrating the RBF score into anomaly detection. Multiplying the RBF layer's dissimilarity score (\( \epsilon_s \)) with the reconstruction error (\( \epsilon_r \)) to form the composite anomaly score (\( \epsilon_s \times \epsilon_r \)) is found to be the most effective, consistently enhancing detection across all benchmarks and metrics. Adding the RBF layer to the vanilla Transformer with only reconstruction error \( \epsilon_r \) as the anomaly score offers marginal improvements on some datasets. In contrast, using only the dissimilarity score (\( \epsilon_s \)) or adding it directly to the reconstruction error (\( \epsilon_s + \epsilon_r \)) shows no significant benefits. 

\begin{figure*}[!t]
\vspace{-1mm}
\centering

\begin{minipage}{0.325\linewidth}
    \includegraphics[width=\linewidth, height=0.8\linewidth]{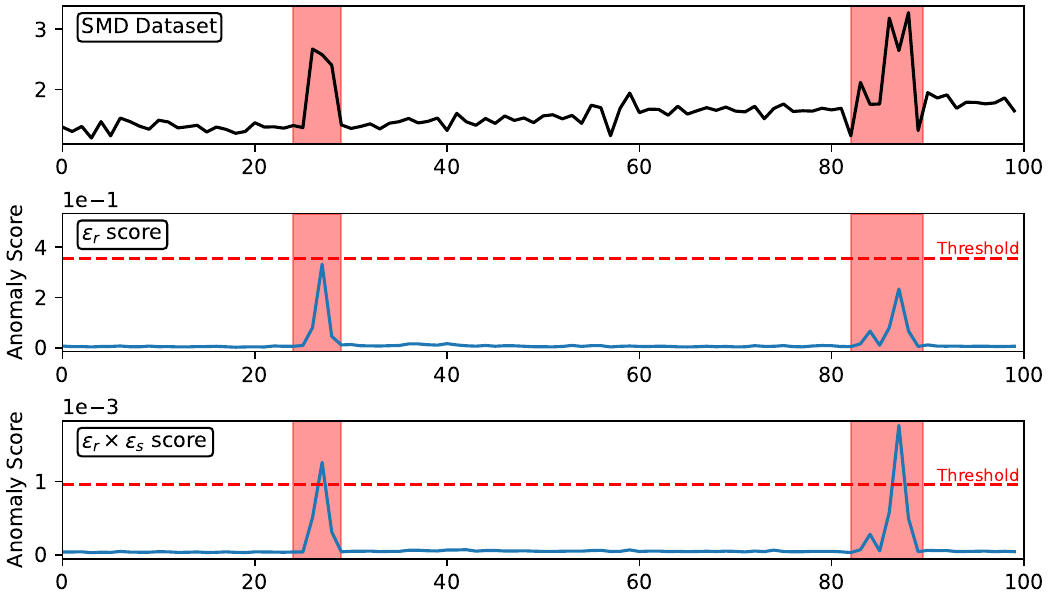}
    \centering
    \vspace{-2mm}
    \textit{(a)}
    \label{fig:SWaT_SOTA_Example}
\end{minipage}\hfill
\begin{minipage}{0.325\linewidth}
    \includegraphics[width=\linewidth, height=0.81\linewidth]{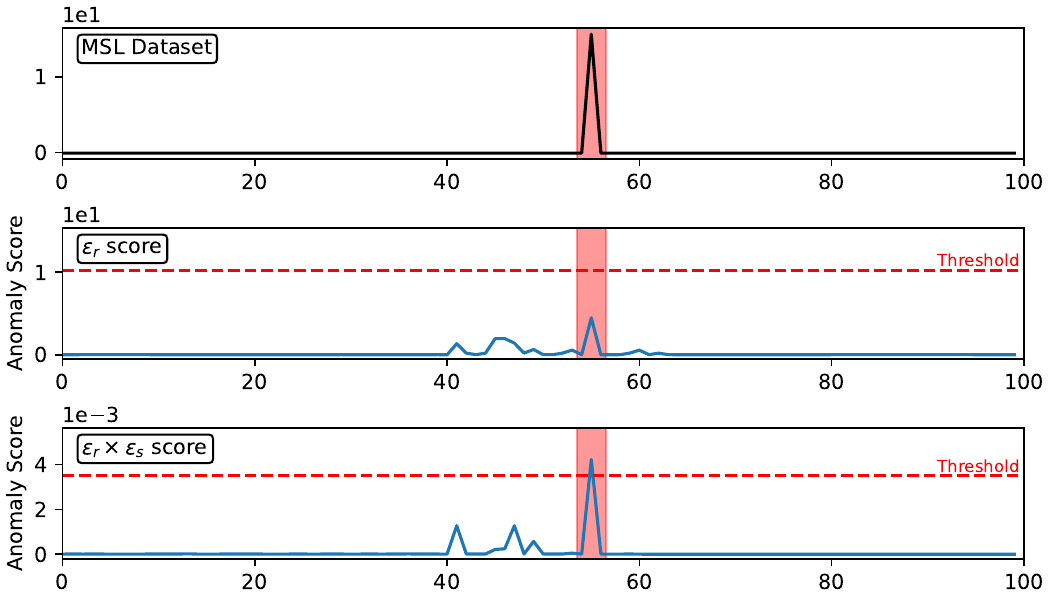} 
    \centering
    \vspace{-2mm}
    \textit{(b)}
    \label{fig:MSL_SOTA_Example}
\end{minipage}\hfill
\begin{minipage}{0.325\linewidth}
    \includegraphics[width=\linewidth, height=0.8\linewidth]{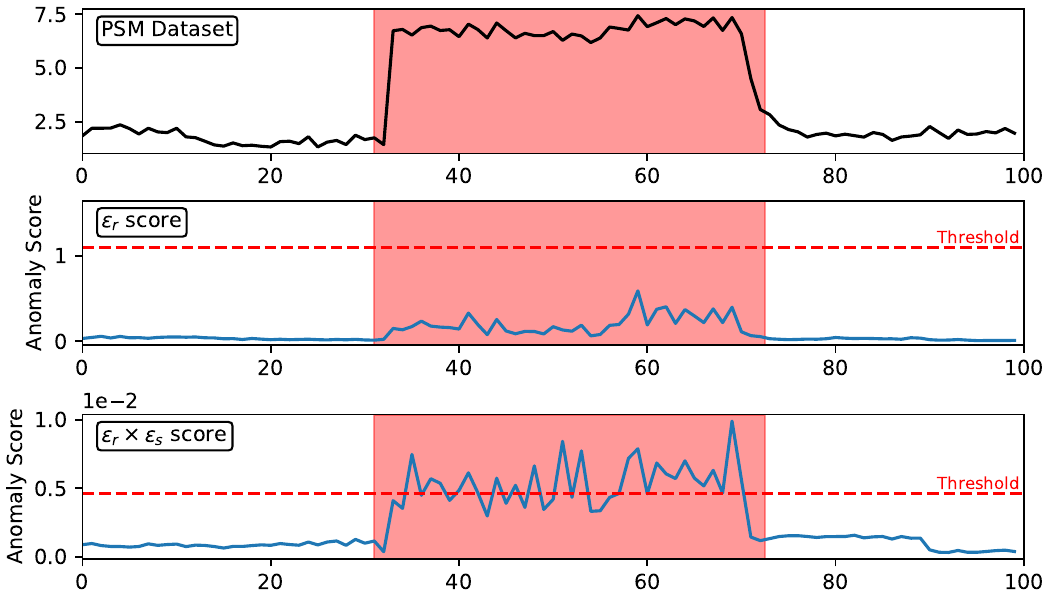}
    \centering
    \vspace{-2mm}
    \textit{(c)}
    \label{fig:SMD_SOTA_Example}
\end{minipage}

\captionof{figure}{\textit{Effect of our composite anomaly score ($\epsilon_r$ $\times$ $\epsilon_s$) compared to reconstruction error ($\epsilon_r$) across segments of all datasets. The highlighted regions in red indicate the true anomaly periods (labeled by an expert).}}
\label{fig:SimRec_Effect}
\vspace{-4mm}
\end{figure*}

Figure \ref{fig:SimRec_Effect} visually illustrates the superiority of our composite anomaly score over the traditional reconstruction score (\( \epsilon_r \)) by showing subsets from all three datasets and the corresponding anomaly scores. Our anomaly score effectively identifies anomalies that are overlooked by the model relying solely on reconstruction error, with detections notably stronger and typically exceeding the threshold. Note that the thresholds depicted in the figures are the best optimized ones based on the entire dataset. Altering this threshold for the subset of data presented in the figures could diminish the overall performance and is therefore not possible.

\

\begin{figure}[!ht]
\centering

\includegraphics[width=0.85\linewidth]{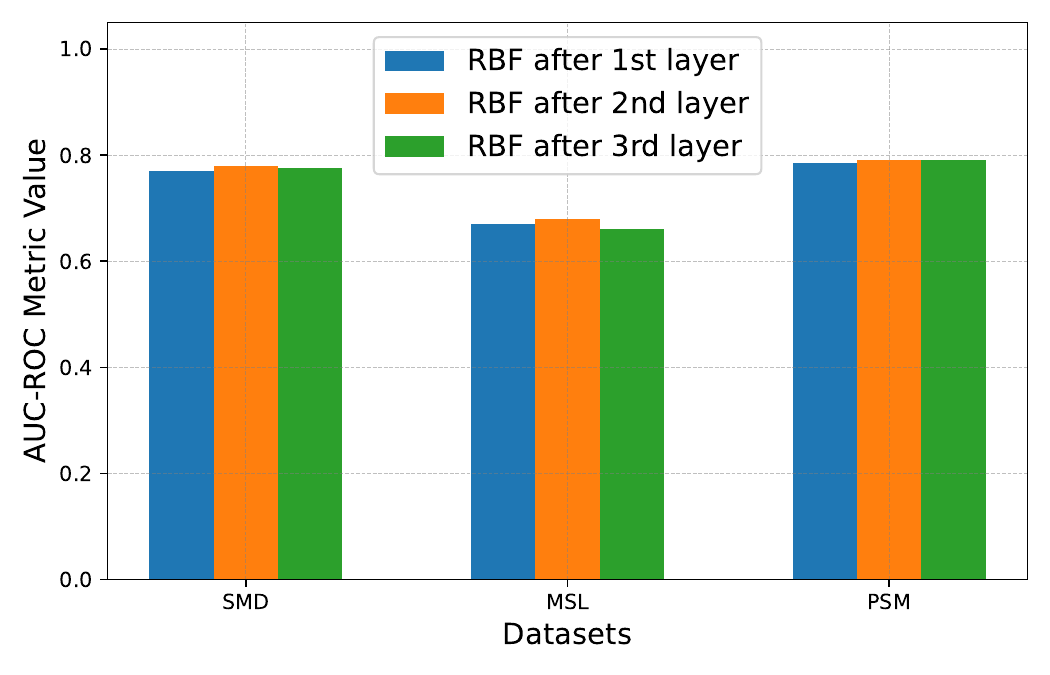}
\vspace{-4mm}
\captionof{figure}{\textit{RESTAD Performance with varying RBF layer placements.}}
\label{fig:RBF_Loc_Random}
\vspace{-5mm}
\end{figure}

\noindent\textbf{RBF Layer Placement}:We explored the flexibility of RBF layer placement within the vanilla Transformer by integrating it after each of the three encoder layers. Figure~\ref{fig:RBF_Loc_Random} demonstrates that performance remains robust across all datasets, irrespective of the RBF layer's location. Note that placing the RBF layer after the second encoder layer results in marginally better performance  across all datasets. This slight advantage influenced our decision to position the RBF layer after the second layer in the final model architecture (see Figure~\ref{Model_Fig}).
\begin{table*}[!t]
    \centering

    \caption{\emph{Effect of integrating RBF layer and the choice of anomaly score. For all measures, a higher value indicates better anomaly detection performance.}}
    \vspace{-1mm}
    \label{tab:Scores_Comparison}
    \fontsize{8pt}{7pt}\selectfont 
    \resizebox{\textwidth}{!}{
    \begin{tabular}{c|c|ccccc|ccccc|ccccc}
    \toprule
    \multirow{6}{*}{Architecture} & \multirow{6}{*}{Anomaly Criterion} &\multicolumn{5}{c}{SMD} & \multicolumn{5}{c}{MSL} & \multicolumn{5}{c}{PSM} \\
    \cmidrule(lr){3-7}\cmidrule(lr){8-12}\cmidrule(lr){13-17}
     & &\rotatebox{90}{F1-Score} & \rotatebox{90}{AUC-ROC} & \rotatebox{90}{AUC-PR} & \rotatebox{90}{VUS-ROC} & \rotatebox{90}{VUS-PR}  &
    \rotatebox{90}{F1-Score} & \rotatebox{90}{AUC-ROC} & \rotatebox{90}{AUC-PR} & \rotatebox{90}{VUS-ROC} & \rotatebox{90}{VUS-PR}  &
    \rotatebox{90}{F1-Score} & \rotatebox{90}{AUC-ROC} & \rotatebox{90}{AUC-PR} & \rotatebox{90}{VUS-ROC} & \rotatebox{90}{VUS-PR} \\
    \midrule
    
    Transformer & $\epsilon_r$ &$0.11$ & $0.75$ & $0.19$ & $0.80$ & $0.22$ 
    & $0.06$ & $0.56$ & $0.14$ & $0.63$ & $0.19$ 
    & $0.13$ & $0.71$ & $0.49$ & $0.70$ & $0.50$ \\
    
    RESTAD & $\epsilon_r$ & $0.11$ & $0.77$ & $0.18$ & $0.81$ & $0.21$ 
    & $0.07$ & $0.63$ & $0.16$ & $0.69$ & $0.21$ 
    & $0.13$ & $0.75$ & $0.56$ & $0.74$ & $0.55$ \\
    
    RESTAD & $\epsilon_s$ & $0.01$ & $0.44$ & $0.03$ & $0.52$ & $0.07$ 
    & $0.01$ & $0.43$ & $0.08$ & $0.48$ & $0.12$ 
    & $0.01$ & $0.32$ & $0.20$ & $0.37$ & $0.25$ \\
    
    RESTAD & $\epsilon_r + \epsilon_s$ & $0.04$ & $0.57$ & $0.06$ & $0.60$ & $0.10$ 
    & $0.07$ & $0.61$ & $0.16$ & $0.65$ & $0.20$ 
    & $0.01$ & $0.68$ & $0.49$ & $0.59$ & $0.45$ \\
    
    \midrule
    \textbf{RESTAD} & $\boldsymbol{\epsilon_r \times \epsilon_s}$ & $\textbf{0.23}$ & $\textbf{0.78}$ & $\textbf{0.23}$ & $\textbf{0.82}$ & $\textbf{0.24}$ 
    & $\textbf{0.07}$ & $\textbf{0.68}$ & $\textbf{0.18}$ & $\textbf{0.72}$ & $\textbf{0.23}$ 
    & $\textbf{0.15}$ & $\textbf{0.79}$ & $\textbf{0.59}$ & $\textbf{0.76}$ & $\textbf{0.57}$ \\

    \bottomrule
    \end{tabular}
    }
\end{table*}

\vspace{-4mm}

\begin{figure}[!h]
\centering

\includegraphics[width=0.85\linewidth]{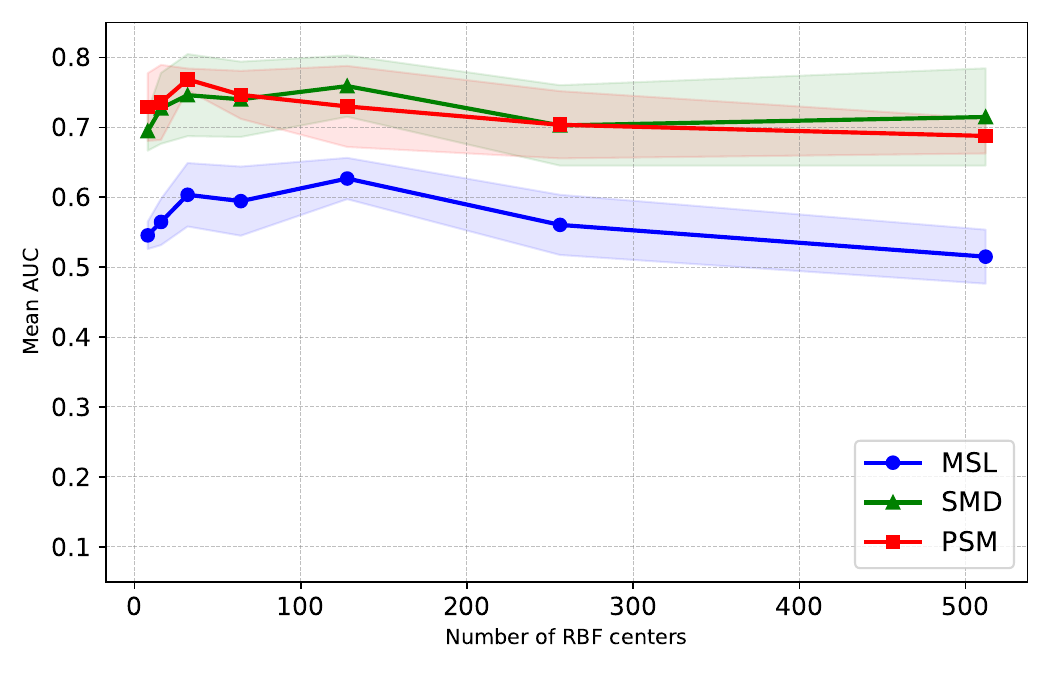}
\vspace{-2mm}
\captionof{figure}{\textit{RESTAD mean performance across varying numbers of RBF centers. Shaded areas indicate ± standard deviation, illustrating variability across multiple runs.}}
\label{fig:RBF_Dim_Random}
\vspace{-6mm}
\end{figure}

\noindent\textbf{Number of RBF Centers}: Figure~\ref{fig:RBF_Dim_Random} represents the impact of the number of centers, ranging from 8 to 512, in the RBF layer of \textit{RESTAD}. Results indicate that the optimal number of RBF centers is data-dependent. Additionally, beyond a certain threshold, increasing the number of centers does not enhance performance and may even reduce it. 

\vspace{-4mm}

\section{Discussion and Conclusion}

We introduced RESTAD, an adaptation of Transformers for unsupervised anomaly detection that improves on the limitations of using only reconstruction error as the anomaly score. By integrating an RBF layer into the Transformer, we combined RBF similarity scores with reconstruction error, enhancing the sensitivity to subtle anomalies. RESTAD consistently outperforms established baselines across various datasets and evaluation metrics.

Our findings reveal that RESTAD’s performance is relatively invariant to RBF layer initialization methods, indicating robustness against initialization variability. The significant performance gains are primarily due to the multiplicative fusion of RBF similarity scores with reconstruction error, markedly improving anomaly detection capabilities. The RBF layer’s placement within the architecture did not significantly affect performance, revealing architectural flexibility in integrating the RBF layer. However, the optimal number of RBF centers is data-dependent. These findings motivate future studies for the exploration of integrating RBF layers into other deep learning architectures for anomaly detection tasks.


\section*{Acknowledgements}
Funding: This work was supported by the Dutch Research Council (NWO) [grant numbers 628.011.214].

{
\fontsize{9pt}{9pt}\selectfont
\bibliography{references}

\begin{thebibliography}{23}
\providecommand{\natexlab}[1]{#1}
\providecommand{\url}[1]{\texttt{#1}}
\expandafter\ifx\csname urlstyle\endcsname\relax
  \providecommand{\doi}[1]{doi: #1}\else
  \providecommand{\doi}{doi: \begingroup \urlstyle{rm}\Url}\fi

\bibitem[Chandola et~al.(2009)Chandola, Banerjee, and Kumar]{Why_Unsupervised1}
Varun Chandola, Arindam Banerjee, and Vipin Kumar.
\newblock Anomaly detection: A survey.
\newblock \emph{ACM computing surveys (CSUR)}, 41\penalty0 (3):\penalty0 1--58, 2009.

\bibitem[Ghorbani et~al.(2024)Ghorbani, Reinders, and Tax]{Why_Unsupervised2}
Ramin Ghorbani, Marcel~JT Reinders, and David~MJ Tax.
\newblock Personalized anomaly detection in ppg data using representation learning and biometric identification.
\newblock \emph{Biomedical Signal Processing and Control}, 94:\penalty0 106216, 2024.

\bibitem[Sch{\"o}lkopf et~al.(1999)Sch{\"o}lkopf, Williamson, Smola, Shawe-Taylor, and Platt]{OC-SVM}
Bernhard Sch{\"o}lkopf, Robert~C Williamson, Alex Smola, John Shawe-Taylor, and John Platt.
\newblock Support vector method for novelty detection.
\newblock \emph{Advances in neural information processing systems}, 12, 1999.

\bibitem[Breunig et~al.(2000)Breunig, Kriegel, Ng, and Sander]{LOF}
Markus~M Breunig, Hans-Peter Kriegel, Raymond~T Ng, and J{\"o}rg Sander.
\newblock Lof: identifying density-based local outliers.
\newblock In \emph{Proceedings of the 2000 ACM SIGMOD international conference on Management of data}, pages 93--104, 2000.

\bibitem[Mejri et~al.(2022)Mejri, Lopez-Fuentes, Roy, Chernakov, Ghorbel, and Aouada]{AD_Limit_with_TS}
Nesryne Mejri, Laura Lopez-Fuentes, Kankana Roy, Pavel Chernakov, Enjie Ghorbel, and Djamila Aouada.
\newblock Unsupervised anomaly detection in time-series: An extensive evaluation and analysis of state-of-the-art methods.
\newblock \emph{arXiv preprint arXiv:2212.03637}, 2022.

\bibitem[Choi et~al.(2021)Choi, Yi, Park, and Yoon]{AD_TS_DeepLearning}
Kukjin Choi, Jihun Yi, Changhwa Park, and Sungroh Yoon.
\newblock Deep learning for anomaly detection in time-series data: review, analysis, and guidelines.
\newblock \emph{IEEE Access}, 9:\penalty0 120043--120065, 2021.

\bibitem[Tuli et~al.(2022)Tuli, Casale, and Jennings]{Transformer_TS}
Shreshth Tuli, Giuliano Casale, and Nicholas~R Jennings.
\newblock Tranad: Deep transformer networks for anomaly detection in multivariate time series data.
\newblock \emph{arXiv preprint arXiv:2201.07284}, 2022.

\bibitem[Vaswani et~al.(2017)Vaswani, Shazeer, Parmar, Uszkoreit, Jones, Gomez, Kaiser, and Polosukhin]{AttentionAllYouNeed}
Ashish Vaswani, Noam Shazeer, Niki Parmar, Jakob Uszkoreit, Llion Jones, Aidan~N Gomez, {\L}ukasz Kaiser, and Illia Polosukhin.
\newblock Attention is all you need.
\newblock \emph{Advances in neural information processing systems}, 30, 2017.

\bibitem[Hundman et~al.(2018)Hundman, Constantinou, Laporte, Colwell, and Soderstrom]{LSTM_Model_Ref}
Kyle Hundman, Valentino Constantinou, Christopher Laporte, Ian Colwell, and Tom Soderstrom.
\newblock Detecting spacecraft anomalies using lstms and nonparametric dynamic thresholding.
\newblock In \emph{Proceedings of the 24th ACM SIGKDD international conference on knowledge discovery \& data mining}, pages 387--395, 2018.

\bibitem[Yang et~al.(2023)Yang, Zhang, Zhou, Wen, and Sun]{DCDetector}
Yiyuan Yang, Chaoli Zhang, Tian Zhou, Qingsong Wen, and Liang Sun.
\newblock Dcdetector: Dual attention contrastive representation learning for time series anomaly detection.
\newblock In \emph{Proceedings of the 29th ACM SIGKDD Conference on Knowledge Discovery and Data Mining}, pages 3033--3045, 2023.

\bibitem[Su et~al.(2019)Su, Zhao, Niu, Liu, Sun, and Pei]{OmnyAnomaly}
Ya~Su, Youjian Zhao, Chenhao Niu, Rong Liu, Wei Sun, and Dan Pei.
\newblock Robust anomaly detection for multivariate time series through stochastic recurrent neural network.
\newblock In \emph{Proceedings of the 25th ACM SIGKDD international conference on knowledge discovery \& data mining}, pages 2828--2837, 2019.

\bibitem[Zhong et~al.(2024{\natexlab{a}})Zhong, Yu, Yang, Wang, and Yang]{PatchAD}
Zhijie Zhong, Zhiwen Yu, Yiyuan Yang, Weizheng Wang, and Kaixiang Yang.
\newblock Patchad: Patch-based mlp-mixer for time series anomaly detection.
\newblock \emph{arXiv preprint arXiv:2401.09793}, 2024{\natexlab{a}}.

\bibitem[Audibert et~al.(2020)Audibert, Michiardi, Guyard, Marti, and Zuluaga]{USAD}
Julien Audibert, Pietro Michiardi, Fr{\'e}d{\'e}ric Guyard, S{\'e}bastien Marti, and Maria~A Zuluaga.
\newblock Usad: Unsupervised anomaly detection on multivariate time series.
\newblock In \emph{Proceedings of the 26th ACM SIGKDD international conference on knowledge discovery \& data mining}, pages 3395--3404, 2020.

\bibitem[Park et~al.(2018)Park, Hoshi, and Kemp]{LSTM-VAE_Paper}
Daehyung Park, Yuuna Hoshi, and Charles~C Kemp.
\newblock A multimodal anomaly detector for robot-assisted feeding using an lstm-based variational autoencoder.
\newblock \emph{IEEE Robotics and Automation Letters}, 3\penalty0 (3):\penalty0 1544--1551, 2018.

\bibitem[Zhao et~al.(2023)Zhao, Jin, Zhou, Li, Liu, and Zhu]{overgeneralization_paper1}
Tianzi Zhao, Liang Jin, Xiaofeng Zhou, Shuai Li, Shurui Liu, and Jiang Zhu.
\newblock Unsupervised anomaly detection approach based on adversarial memory autoencoders for multivariate time series.
\newblock \emph{Computers, Materials \& Continua}, 76\penalty0 (1), 2023.

\bibitem[Zhong et~al.(2024{\natexlab{b}})Zhong, Zhao, and Lim]{overgeneralization_paper2}
Haoyi Zhong, Yongjiang Zhao, and Chang~Gyoon Lim.
\newblock Abnormal state detection using memory-augmented autoencoder technique in frequency-time domain.
\newblock \emph{KSII Transactions on Internet and Information Systems (TIIS)}, 18\penalty0 (2):\penalty0 348--369, 2024{\natexlab{b}}.

\bibitem[Xu et~al.(2021)Xu, Wu, Wang, and Long]{AnomalyTransformer}
Jiehui Xu, Haixu Wu, Jianmin Wang, and Mingsheng Long.
\newblock Anomaly transformer: Time series anomaly detection with association discrepancy.
\newblock \emph{arXiv preprint arXiv:2110.02642}, 2021.

\bibitem[Orr et~al.(1996)]{RBF_Kernel}
Mark~JL Orr et~al.
\newblock Introduction to radial basis function networks, 1996.

\bibitem[Abdulaal et~al.(2021)Abdulaal, Liu, and Lancewicki]{PSM_Dataset}
Ahmed Abdulaal, Zhuanghua Liu, and Tomer Lancewicki.
\newblock Practical approach to asynchronous multivariate time series anomaly detection and localization.
\newblock In \emph{Proceedings of the 27th ACM SIGKDD conference on knowledge discovery \& data mining}, pages 2485--2494, 2021.

\bibitem[Shen et~al.(2020)Shen, Li, and Kwok]{Sliding_Window_Method}
Lifeng Shen, Zhuocong Li, and James Kwok.
\newblock Timeseries anomaly detection using temporal hierarchical one-class network.
\newblock \emph{Advances in Neural Information Processing Systems}, 33:\penalty0 13016--13026, 2020.

\bibitem[Paparrizos et~al.(2022)Paparrizos, Boniol, Palpanas, Tsay, Elmore, and Franklin]{VUS_Evaluation}
John Paparrizos, Paul Boniol, Themis Palpanas, Ruey~S Tsay, Aaron Elmore, and Michael~J Franklin.
\newblock Volume under the surface: a new accuracy evaluation measure for time-series anomaly detection.
\newblock \emph{Proceedings of the VLDB Endowment}, 15\penalty0 (11):\penalty0 2774--2787, 2022.

\bibitem[Xu et~al.(2018)Xu, Chen, Zhao, Li, Bu, Li, Liu, Zhao, Pei, Feng, et~al.]{PA_1}
Haowen Xu, Wenxiao Chen, Nengwen Zhao, Zeyan Li, Jiahao Bu, Zhihan Li, Ying Liu, Youjian Zhao, Dan Pei, Yang Feng, et~al.
\newblock Unsupervised anomaly detection via variational auto-encoder for seasonal kpis in web applications.
\newblock In \emph{Proceedings of the 2018 world wide web conference}, pages 187--196, 2018.

\bibitem[Kim et~al.(2022)Kim, Choi, Choi, Lee, and Yoon]{PA_K}
Siwon Kim, Kukjin Choi, Hyun-Soo Choi, Byunghan Lee, and Sungroh Yoon.
\newblock Towards a rigorous evaluation of time-series anomaly detection.
\newblock In \emph{Proceedings of the AAAI Conference on Artificial Intelligence}, volume~36, pages 7194--7201, 2022.

\end{thebibliography}
}


\end{document}